\documentclass[11pt]{article}
\usepackage{emnlp2023}
\usepackage{tcolorbox}
\usepackage{latexsym}
\usepackage{graphicx}
\usepackage{amsmath}
\usepackage{times}
\usepackage{pgfplots}
\usepackage{pgfplotstable}
\usepackage{booktabs}
\usepackage{pifont}
\pgfplotsset{compat=1.18}

\newcommand{\MinMax}[3]{%
  ( (#3) - (#2) < 1e-12 ? 0.5 : ( (#1) - (#2) ) / ((#3) - (#2)) )%
}

\newcommand{\MinMaxFlip}[3]{%
  1 - \MinMax{#1}{#2}{#3}%
}

\newcommand{\BoldIfEqFmt}[1]{%
  \pgfkeysgetvalue{/pgfplots/table/@cell content}\celltxt%
  \edef\celltxt{\celltxt}%
  \edef\maxtxt{#1}%
  \ifx\celltxt\maxtxt
    \pgfkeysalso{/pgfplots/table/@cell content/.add={\boldmath }{}}%
  \fi
}

\newcommand{\SmartTableNine}[3]{%
  \begingroup

  \pgfplotstableread[col sep=semicolon, row sep=\\]{%
Str;ROUGE;METEOR;BERTScore;FKGL;DCRS;SLE;SummaC;FENICE;CHEX\\
MIN;#2\\
MAX;#3\\
#1
  }\datatable

  \pgfplotstablegetelem{0}{ROUGE}\of{\datatable}\let\ROUGEmin\pgfplotsretval
  \pgfplotstablegetelem{1}{ROUGE}\of{\datatable}\let\ROUGEmax\pgfplotsretval

  \pgfplotstablegetelem{0}{METEOR}\of{\datatable}\let\METEORmin\pgfplotsretval
  \pgfplotstablegetelem{1}{METEOR}\of{\datatable}\let\METEORmax\pgfplotsretval

  \pgfplotstablegetelem{0}{BERTScore}\of{\datatable}\let\BERTmin\pgfplotsretval
  \pgfplotstablegetelem{1}{BERTScore}\of{\datatable}\let\BERTmax\pgfplotsretval

  \pgfplotstablegetelem{0}{FKGL}\of{\datatable}\let\FKGLmin\pgfplotsretval
  \pgfplotstablegetelem{1}{FKGL}\of{\datatable}\let\FKGLmax\pgfplotsretval

  \pgfplotstablegetelem{0}{DCRS}\of{\datatable}\let\DCRSmin\pgfplotsretval
  \pgfplotstablegetelem{1}{DCRS}\of{\datatable}\let\DCRSmax\pgfplotsretval

  \pgfplotstablegetelem{0}{SLE}\of{\datatable}\let\SLEmin\pgfplotsretval
  \pgfplotstablegetelem{1}{SLE}\of{\datatable}\let\SLEmax\pgfplotsretval

  \pgfplotstablegetelem{0}{SummaC}\of{\datatable}\let\SUMMACmin\pgfplotsretval
  \pgfplotstablegetelem{1}{SummaC}\of{\datatable}\let\SUMMACmax\pgfplotsretval

  \pgfplotstablegetelem{0}{FENICE}\of{\datatable}\let\FENICEmin\pgfplotsretval
  \pgfplotstablegetelem{1}{FENICE}\of{\datatable}\let\FENICEmax\pgfplotsretval

  \pgfplotstablegetelem{0}{CHEX}\of{\datatable}\let\OPENFACTmin\pgfplotsretval
  \pgfplotstablegetelem{1}{CHEX}\of{\datatable}\let\OPENFACTmax\pgfplotsretval

  \pgfmathprintnumberto[fixed,precision=4]{\ROUGEmax}{\ROUGEmaxfmt}
  \pgfmathprintnumberto[fixed,precision=4]{\METEORmax}{\METEORmaxfmt}
  \pgfmathprintnumberto[fixed,precision=4]{\BERTmax}{\BERTmaxfmt}
  \pgfmathprintnumberto[fixed,precision=4]{\SUMMACmax}{\SUMMACmaxfmt}
  \pgfmathprintnumberto[fixed,precision=4]{\FENICEmax}{\FENICEmaxfmt}
  \pgfmathprintnumberto[fixed,precision=4]{\OPENFACTmax}{\OPENFACTmaxfmt}

  \pgfmathprintnumberto[fixed,precision=2]{\FKGLmin}{\FKGLminfmt}
  \pgfmathprintnumberto[fixed,precision=4]{\DCRSmin}{\DCRSminfmt}
  \pgfmathprintnumberto[fixed,precision=4]{\SLEmin}{\SLEminfmt}

  \pgfplotstablecreatecol[
    create col/expr={\MinMax{\thisrow{ROUGE}}{\ROUGEmin}{\ROUGEmax}}
  ]{ROUGE_N}{\datatable}

  \pgfplotstablecreatecol[
    create col/expr={\MinMax{\thisrow{METEOR}}{\METEORmin}{\METEORmax}}
  ]{METEOR_N}{\datatable}

  \pgfplotstablecreatecol[
    create col/expr={\MinMax{\thisrow{BERTScore}}{\BERTmin}{\BERTmax}}
  ]{BERT_N}{\datatable}

  \pgfplotstablecreatecol[
    create col/expr={\MinMaxFlip{\thisrow{FKGL}}{\FKGLmin}{\FKGLmax}}
  ]{FKGL_N}{\datatable}

  \pgfplotstablecreatecol[
    create col/expr={\MinMaxFlip{\thisrow{DCRS}}{\DCRSmin}{\DCRSmax}}
  ]{DCRS_N}{\datatable}

  \pgfplotstablecreatecol[
    create col/expr={\MinMaxFlip{\thisrow{SLE}}{\SLEmin}{\SLEmax}}
  ]{SLE_N}{\datatable}

  \pgfplotstablecreatecol[
    create col/expr={\MinMax{\thisrow{SummaC}}{\SUMMACmin}{\SUMMACmax}}
  ]{SUMMAC_N}{\datatable}

  \pgfplotstablecreatecol[
    create col/expr={\MinMax{\thisrow{FENICE}}{\FENICEmin}{\FENICEmax}}
  ]{FENICE_N}{\datatable}

  \pgfplotstablecreatecol[
    create col/expr={\MinMax{\thisrow{CHEX}}{\OPENFACTmin}{\OPENFACTmax}}
  ]{OPENFACT_N}{\datatable}

  \pgfplotstablecreatecol[
    create col/expr={
      (
        \thisrow{ROUGE_N}+
        \thisrow{METEOR_N}+
        \thisrow{BERT_N}+
        \thisrow{FKGL_N}+
        \thisrow{DCRS_N}+
        \thisrow{SLE_N}+
        \thisrow{SUMMAC_N}+
        \thisrow{FENICE_N}+
        \thisrow{OPENFACT_N}
      ) / 9
    }
  ]{Mean}{\datatable}

  \pgfplotstabletypeset[
    row predicate/.code={%
      \pgfplotstablegetelem{\pgfplotstablerow}{Str}\of{\datatable}%
      \edef\cell{\pgfplotsretval}%
      \def\MINROW{MIN}%
      \def\MAXROW{MAX}%
      \ifx\cell\MINROW\pgfplotstableuserowfalse\fi
      \ifx\cell\MAXROW\pgfplotstableuserowfalse\fi
    },
    columns={
      Str,
      ROUGE,
      METEOR,
      BERTScore,
      FKGL,
      DCRS,
      SLE,
      SummaC,
      FENICE,
      CHEX,
      Mean
    },
    columns/Str/.style={
      string type,
      column name=Str
    },
    columns/ROUGE/.style={
      column name=ROUGE$\uparrow$,
      fixed,
      precision=4,
      postproc cell content/.append code={\BoldIfEqFmt{\ROUGEmaxfmt}}
    },
    columns/METEOR/.style={
      column name=METEOR$\uparrow$,
      fixed,
      precision=4,
      postproc cell content/.append code={\BoldIfEqFmt{\METEORmaxfmt}}
    },
    columns/BERTScore/.style={
      column name=BERTScore$\uparrow$,
      fixed,
      precision=4,
      postproc cell content/.append code={\BoldIfEqFmt{\BERTmaxfmt}}
    },
    columns/FKGL/.style={
      column name=FKGL$\downarrow$,
      fixed,
      precision=2,
      postproc cell content/.append code={\BoldIfEqFmt{\FKGLminfmt}}
    },
    columns/DCRS/.style={
      column name=DCRS$\downarrow$,
      fixed,
      precision=4,
      postproc cell content/.append code={\BoldIfEqFmt{\DCRSminfmt}}
    },
    columns/SLE/.style={
      column name=SLE$\downarrow$,
      fixed,
      precision=4,
      postproc cell content/.append code={\BoldIfEqFmt{\SLEminfmt}}
    },
    columns/SummaC/.style={
      column name=SummaC$\uparrow$,
      fixed,
      precision=4,
      postproc cell content/.append code={\BoldIfEqFmt{\SUMMACmaxfmt}}
    },
    columns/FENICE/.style={
      column name=FENICE$\uparrow$,
      fixed,
      precision=4,
      postproc cell content/.append code={\BoldIfEqFmt{\FENICEmaxfmt}}
    },
    columns/CHEX/.style={
      column name=CHEX$\uparrow$,
      fixed,
      precision=4,
      postproc cell content/.append code={\BoldIfEqFmt{\OPENFACTmaxfmt}}
    },
    columns/Mean/.style={
      column name=Mean$\uparrow$,
      fixed,
      precision=4
    },
    every head row/.style={
      before row={
        \toprule
        & \multicolumn{3}{c}{Relevance}
        & \multicolumn{3}{c}{Readability}
        & \multicolumn{3}{c}{Factuality}
        & \\[-0.8ex]
        \cmidrule(lr){2-4}
        \cmidrule(lr){5-7}
        \cmidrule(lr){8-10}
      },
      after row=\midrule
    },
    every last row/.style={
      after row=\bottomrule
    },
    font=\normalsize
   ]{\datatable}
  \endgroup
}

\title{Improving Health Literacy through Lay Summarization of Radiological Reports: An Evaluation of BioNER and Retrieval-Augmented Generation}

\author{
Egecan Çelik Evgin\textsuperscript{1},
İlknur Karadeniz\textsuperscript{3},
Olcay Taner Yıldız\textsuperscript{1,2} \\
\textsuperscript{1}Department of Artificial Intelligence and Data Engineering, Özyeğin University, Türkiye \\
\textsuperscript{2}Department of Computer Science, Özyeğin University, Türkiye \\
\textsuperscript{3}Department of Computer Engineering, Galatasaray University, Türkiye \\
\texttt{egecan.evgin@ozu.edu.tr}, \\
\texttt{ikaradeniz@gsu.edu.tr}, \\
\texttt{olcay.yildiz@ozyegin.edu.tr}
}

\begin{document}

\maketitle

\begin{abstract}

Radiology reports are written primarily for clinicians, and their specialized terminology often makes them difficult for patients to interpret. As a result, many patients turn to publicly available Large Language Models (LLMs) to help explain their reports, despite well-documented risks of factual inaccuracies and hallucinations. Automated lay-summary generation has emerged as a promising alternative, yet the effectiveness of retrieval-enhanced and clinically informed approaches for radiology-specific communication remains underexplored. This study investigates the extent to which Retrieval-Augmented Generation (RAG) and Named Entity Recognition (NER) improve the quality, factual consistency, and readability of automatically generated lay summaries compared with standard LLM-based generation. We develop a framework combining NER-based extraction of clinically relevant findings with a RAG mechanism for contextual grounding, evaluated across few-shot and fine-tuned variants of two models (Qwen, BioBART). Results show that NER consistently improves readability and overall quality, while RAG alone offers no benefit and can introduce hallucinations from irrelevant retrieved terms. Combining RAG with NER degrades performance in few-shot settings but improves readability when fine-tuned. Fine-tuned BioBART with NER achieves the best overall performance, highlighting entity-aware extraction as the primary driver of improved patient-friendly summaries.

\end{abstract}

\section{Introduction}

Radiological reports document the findings of medical imaging examinations, such as X-rays, computed tomography (CT), and magnetic resonance imaging (MRI), and serve as a primary means of communication between healthcare professionals. However, these reports are typically written using specialized biomedical terminology and complex clinical language, making them difficult for patients to understand. Consequently, many patients struggle to interpret their imaging results and fully comprehend the implications of the reported findings.
To better understand their medical conditions and make informed decisions about treatment, patients often seek additional information. Traditionally, this involved searching online for medical terms and symptoms. Today, many patients use Large Language Model (LLM)-based chatbots, such as ChatGPT, Gemini, and DeepSeek, to obtain health-related information \cite{openai2022chatgpt,google2024gemini,deepseek2024v3}. However, these systems can generate inaccurate or hallucinated content, potentially leading patients to misunderstand their radiological findings or place undue trust in incorrect information.

This communication gap can limit patient understanding and health literacy, motivating research into methods that translate radiological reports into patient-friendly language. Recent work has explored Retrieval-Augmented Generation (RAG) to improve the quality and factual consistency of lay summarization by grounding generated outputs in external knowledge \cite{guo2024retrieval}. In parallel, Named Entity Recognition (NER) has been used to identify clinically relevant terms that can guide and constrain generation toward more accurate and relevant content. However, the comparative effectiveness of retrieval-based and entity-aware approaches for radiology-specific lay summarization remains underexplored, particularly across models of different scale and training regime.

To address this gap, we evaluate our approach across four public radiology report datasets spanning diverse clinical settings and imaging modalities: PadChest, BIMCV-COVID19+, Open-i, and MIMIC-CXR \cite{bustos2020padchest,vaya2020bimcv,demner2012openi,johnson2019mimiccxr}.

The main contributions of this study are:

\begin{itemize}
    \item A framework combining and comparing RAG-based and NER-enhanced approaches to radiology report lay summarization;
    \item A comparison between a state-of-the-art general-purpose LLM and a biomedical small language model;
    \item The use of few-shot baselines to systematically compare against fine-tuned model variants.
\end{itemize}

The remainder of the paper is organized as follows: Section 2 reviews related work, Section 3 describes the methodology, Section 4 presents the results and discussion, and Section 5 concludes the paper.

\section{Related Work}

Lay summaries differ from standard summaries in their emphasis on readability for non-expert audiences. In the biomedical domain, the BioLaySumm shared task has been organized in 2023, 2024, and 2025 \cite{goldsack-etal-2023-biolaysumm,goldsack-etal-2024-overview,xiao-etal-2025-overview}, aiming to generate lay summaries that are relevant, readable, and factual. BioLaySumm 2025 Shared Task 2 focused specifically on generating lay summaries from radiology reports, and several of the approaches discussed below were developed for this task.

\textbf{Fine-tuning-based approaches:} AEHRC achieved the best overall performance in both the open and closed subtasks of Shared Task 2 using fully supervised fine-tuning, comparing T5-Large with LLaMA-3.2-3B and finding T5-Large superior, without using LoRA, quantization, or RAG \cite{zhang-etal-2025-aehrc,raffel2020exploring,meta2024llama32}. KHU\_LDI, the second-best open-track system, used QLoRA fine-tuning on Qwen2.5-3B-Instruct and Qwen3-4B, combined with 3-shot prompting and a generate-feedback-refine pipeline \cite{moriazi-sung-2025-khu-ldi,dettmers2023qlora,yang2024qwen25,yang2025qwen3,madaan2023selfrefine}. MetninOzU ranked third overall using an abstract-based summarization setup, showing that shorter inputs can still yield strong factuality scores \cite{evgin-etal-2025-metninozu}.

\textbf{Prompting-based approaches:} 5cNLP, the second-place closed-track system, relied on structured prompting rather than fine-tuning, testing Llama-3.3-70B-Instruct and GPT-4.1; their best result used GPT-4.1 with few-shot radiology examples selected via BERT-large embeddings \cite{lossio-ventura-etal-2025-5cnlp}. Proff et al. compared GPT-4o, Llama-3-70B, and Mixtral-8x22B for radiology report simplification, finding that all models improved readability, though open-weight models produced more high-risk errors than GPT-4o \cite{proff2026simplifying}.

\textbf{RAG-based approaches:} CUTN\_Bio placed third in the closed track of Shared Task 2 using a RAG pipeline with Zephyr-7B-beta, extracting medical terms via SciSpacy and retrieving Wikipedia definitions stored in ChromaDB \cite{sivagnanam-etal-2025-cutn,tunstall2023zephyr,neumann-etal-2019-scispacy,chroma2026}. The same team placed second in Subtask 1.2 (external-knowledge lay summarization) using a similar RAG approach with MedCAT for term extraction and LLaMA-3-8B-Instruct for generation \cite{kraljevic2021medcat,meta2024llama32}. Sun et al. proposed FactMM-RAG, which retrieves factually similar report content via RadGraph prior to generation to improve the accuracy of generated radiology reports \cite{sun-etal-2025-fact}. LaySummX applied retrieval-augmented fine-tuning, using abstracts to retrieve relevant full-text chunks before fine-tuning LLaMA 3.1 with LoRA \cite{lin-yu-2025-laysummx}. Guo et al. introduced Retrieval-Augmented Lay Language generation, retrieving UMLS and Wikipedia definitions to supply missing background explanations \cite{guo2024retrieval}. UIUC\_BioNLP used an extract-then-summarize pipeline combining Wikipedia definition retrieval with DPR-based passage retrieval \cite{you-etal-2024-uiuc}.

\textbf{NER-based approaches:} ISIKSumm used a BART-based system augmented with biomedical entity labels using Stanza NER to improve handling of technical terms \cite{colak-karadeniz-2023-isiksumm}. Gupta and Krishnamurthy's LayForge system used BioBERT NER to identify biomedical terms and incorporated UMLS definitions before summary rewriting, improving readability and factuality at a small cost to ROUGE scores \cite{gupta-krishnamurthy-2025-shared,lee2020biobert,bodenreider2004umls,lin-2004-rouge}. Ming et al. used MeSH terms to guide LLMs toward more informative background context for lay readers \cite{ming-etal-2025-towards}.

Overall, prior work has largely explored RAG-based and NER-based strategies in isolation, with few studies directly comparing their effectiveness within a unified framework, or across models differing substantially in scale and training regime (few-shot vs. fine-tuned). This study addresses this gap by systematically comparing RAG-based and NER-enhanced lay summarization strategies for radiology reports.

\section{Methodology}

\subsection{Models}

Two models were used in this study: Qwen3.5-0.8B \cite{qwen35}, a recent general-purpose small language model, and BioBART-v2-large \cite{BioBART}, a model pretrained specifically on biomedical text. This pairing enables comparison between a strong general-purpose model and a smaller model adapted for the biomedical domain.

\subsection{Dataset and Evaluation Set}

Four public radiology report datasets, spanning diverse clinical settings and imaging modalities, were used in this study: PadChest, BIMCV-COVID19+, Open-i, and MIMIC-CXR. PadChest contains over 160,000 chest X-ray images from approximately 67,000 patients, while BIMCV-COVID19+ includes COVID-19 X-ray and CT studies, comprising 21,342 CR, 34,829 DX, and 7,918 CT cases \cite{bustos2020padchest,vaya2020bimcv}. Open-i is smaller, with 7,470 chest X-ray images and 3,955 reports, while MIMIC-CXR is the largest dataset, with 227,835 studies and 377,110 images derived from real clinical reports \cite{demner2012openi,johnson2019mimiccxr}.
Lay summaries were automatically generated from the clinical reports using the Layman's RRG framework \cite{zhao-etal-2026-x}, and the combined datasets were used in the BioLaySumm 2025 shared task \cite{xiao-etal-2025-overview}. However, the lay summaries for the shared task's test set were not publicly available. To address this, the original training set was split to construct a new test set, with the goal of obtaining a larger evaluation set than the one used in the shared task. The split was performed randomly (seed = 42) to avoid bias; this is distinct from the sampling of the three few-shot exemplar reports described in §3.4, which used the same seed value for a separate sampling step.
The resulting split comprises 168,036 training reports (89.38\%), 14,971 validation reports (7.96\%), 5,000 test reports (2.66\%), and 3 few-shot exemplar reports, with average token counts summarized in Table~\ref{tab:dataset_split_stats}.

\begin{table}[ht]
\centering
\caption{Dataset split statistics}
\label{tab:dataset_split_stats}
\resizebox{\columnwidth}{!}{%
\begin{tabular}{lrrrr}
\hline
\textbf{Split Name} & \textbf{Split \%} & \textbf{Rows} & \textbf{Rad. Reports} & \textbf{Lay} \\
\hline
Training & 89.38\% & 168,036 & 31.11 & 42.17 \\
Validation & 7.96\% & 14,971 & 34.27 & 45.49 \\
Testing & 2.66\% & 5,000 & 31.19 & 42.22 \\
Few-Shot Samples & 0.001\% & 3 & 11.00 & 20.33 \\
\hline
\end{tabular}%
}
\vspace{1mm}
\footnotesize{Rad. Reports: Average tokens in radiological reports;\\ 
Lay: Average tokens in layman summaries.}
\end{table}

\subsection{Evaluation Metrics}

Generated summaries were evaluated along three dimensions: relevance, readability, and factuality. All metrics were scaled using min-max normalization to place their values on a comparable range \cite{han2011datamining}. No additional weighting was applied across the three metric groups, since each group contained an equal number of metrics. For relevance and factuality metrics, higher scores indicate better performance; for all readability metrics (FKGL, DCRS, SLE), lower scores indicate better performance (i.e., simpler, more accessible text).

\textbf{Relevance:} ROUGE \cite{lin-2004-rouge} measures word overlap between predicted and gold lay summaries; we report the average F1 across ROUGE-1, ROUGE-2, and ROUGE-L. METEOR \cite{banerjee-lavie-2005-meteor} extends beyond exact word matches by accounting for stems and synonyms, offering a complementary view of relevance. BERTScore \cite{zhang2020bertscore} measures semantic similarity by comparing contextual word embeddings between predicted and gold summaries, computing precision, recall, and F1 based on closest token matches.

\textbf{Readability:} FKGL \cite{kincaid1975readability} estimates the grade level of a summary based on sentence and word length, with longer sentences and words yielding higher (less readable) scores. DCRS \cite{dale-chall-1948-readability} complements FKGL by assessing word familiarity against a list of common words, capturing cases FKGL may miss, such as short but unfamiliar words (e.g., "understand"). SLE \cite{cripwell-etal-2023-simplicity} is a transformer-based metric that requires only the predicted summary, using a RoBERTa-base model with a regression head to produce a simplicity score.

\textbf{Factuality:} SummaC \cite{laban-etal-2022-summac} evaluates sentence-level agreement between the source report and the predicted summary using entailment and contradiction scores, penalizing contradictory content. FENICE \cite{scire-etal-2024-fenice} evaluates factuality at the claim level by extracting atomic claims from the predicted summary and verifying them against the source text, directly penalizing unsupported or contradicted claims. CheXbert-F1 \cite{smit-etal-2020-chexbert}, developed specifically for radiology reports, evaluates whether clinical findings in the generated summary match those in the reference report, penalizing missing or incorrect findings.

\subsection{Baseline Strategies}

Two baseline strategies were established for each model:
(i) Few-shot prompting. Baselines were computed under 0-shot, 1-shot, and 3-shot settings. To construct the 1-shot and 3-shot exemplars, three radiology reports were randomly sampled (seed = 42) and excluded from the test set to prevent data leakage (Table~\ref{tab:dataset_split_stats}).
(ii) LoRA fine-tuning. 

Both models were fine-tuned using LoRA with r=4, \texttt{lora\_alpha=8}, dropout of 0.05, and no bias term. Qwen was adapted on \texttt{q\_proj} and \texttt{v\_proj}; BioBART was adapted on \texttt{q\_proj}, \texttt{v\_proj}, and \texttt{out\_proj}. Training used 2 epochs, a batch size of 20 (evaluation batch size 16), gradient accumulation of 1, learning rate \texttt{2e-4}, weight decay \texttt{0.01}, 100 warmup steps, with \texttt{bf16} enabled and \texttt{fp16} disabled.

\subsection{Enhancement Strategies}

In addition to the baselines, three enhancement strategies were evaluated, each applied to both the few-shot and fine-tuned settings of both models: \begin{itemize}
\item NER-enhanced (BioNER): Clinically relevant terms were extracted using Stanza's radiology NER model \cite{zhang2021biomedical}, which identifies five entity classes: ANATOMY, OBSERVATION, ANATOMY\_MODIFIER, OBSERVATION\_MODIFIER, and UNCERTAINTY. Extracted entities were used to guide summary generation toward clinically relevant content.
\item RAG-enhanced: A retrieval-augmented pipeline was built in which an agent extracted candidate medical terms from the source report. Each term was first checked against a local term-description database; if not found, it was searched via the Wikipedia API, and the first sentence of the result was stored in the local database for reuse. Retrieved definitions were then provided as contextual grounding during summary generation.
\item Combined (BioNER + RAG): Term extraction was performed using BioNER, and the resulting terms were used to query the RAG retrieval pipeline described above, combining entity-guided extraction with retrieval-based grounding.
\end{itemize}
This produces three conditions per model per learning setting (baseline, +BioNER, +RAG, +BioNER+RAG).

\section{Results} 

Table~\ref{tab:mean_scores_few_shot_strategies} presents the overall results of the few-shot strategies. For the Qwen model, the BioNER strategy improved overall relevance, readability, and factuality compared with the 0-shot baseline. The RAG strategy did not outperform the baseline, while the combination of BioNER and RAG resulted in lower scores across all evaluation dimensions. A similar trend was observed for BioBART. Compared with Qwen, BioBART achieved lower performance in the few-shot setting across most evaluation metrics.

Table~\ref{tab:mean_scores_fine_tuning_strategies} summarizes the fine-tuning results. In contrast to the few-shot experiments, BioBART outperformed Qwen after fine-tuning. Similar to the few-shot setting, the BioNER strategy improved overall performance, particularly readability, for both models. The RAG strategy improved readability but reduced relevance in both models. For Qwen, the combined BioNER+RAG strategy achieved the best FKGL and DCRS scores together with the lowest SLE score.

Table~\ref{tab:mean_scores_best_ft_few_shot_each_model} compares the best-performing few-shot and fine-tuning strategies for each model. For Qwen, the best few-shot strategy (0-shot BioNER) outperformed all fine-tuning configurations. In contrast, BioBART achieved its highest performance after fine-tuning. Comparing the best-performing configurations of both models, fine-tuned BioBART with BioNER achieved better overall performance than Qwen with the 0-shot BioNER strategy across most evaluation metrics.

\begin{table*}[ht]
\centering
\caption{Mean scores for Few-Shot Based Strategies}
\label{tab:mean_scores_few_shot_strategies}
\resizebox{\textwidth}{!}{%
\SmartTableNine{
Qwen3.5 0-Shot;0.357039;0.395199;0.914621;6.594260;9.809669;1.490906;0.578389;0.320894;0.891837\\
Qwen3.5 1-Shot;0.327155;0.358100;0.907675;6.713330;9.797644;1.376298;0.656704;0.203194;0.897719\\
Qwen3.5 3-Shot;0.368775;0.402880;0.914017;6.257193;9.305661;1.559115;0.745062;0.401754;0.905556\\
Qwen3.5 BioNER 0-Shot;0.358850;0.399018;0.918335;6.347393;9.384134;1.068674;0.602776;0.412674;0.894134\\
Qwen3.5 RAG 0-Shot;0.356489;0.389763;0.908850;7.098400;9.639643;1.310610;0.585504;0.174225;0.888963\\
Qwen3.5 BioNER + RAG 0-Shot;0.247364;0.301855;0.864979;9.610888;10.742594;1.854264;0.541981;0.050681;0.814128\\
BioBART 0-Shot;0.194001;0.280669;0.844770;15.386576;11.901612;1.044979;0.300042;0;0.770808\\
BioBART 1-Shot;0.084542;0.129753;0.813567;15.201627;12.866095;0.529589;0.417323;0.334057;0.107408\\
BioBART 3-Shot;0.087234;0.141804;0.813700;14.455682;13.071259;0.540807;0.582867;0.349791;0.036848\\
BioBART BioNER 0-Shot;0.089872;0.124094;0.802036;18.389608;12.137616;0.656499;0.415961;0.384148;0.753100\\
BioBART RAG 0-Shot;0.106918;0.165445;0.818577;14.313906;11.831459;1.313984;0.626585;0.032204;0.768604\\
BioBART BioNER + RAG 0-Shot;0.091773;0.135059;0.805173;14.662896;11.286561;0.984897;0.503310;0.144237;0.777952\\
}{
0.084542;0.124094;0.802036;6.257193;9.305661;0.529589;0.300042;0;0.036848
}{
0.368775;0.402880;0.918335;18.389608;13.071259;1.854264;0.745062;0.412674;0.905556
}%
}
\end{table*}

\begin{table*}[ht]
\centering
\caption{Mean scores for Fine-Tuning Based Strategies}
\label{tab:mean_scores_fine_tuning_strategies}
\resizebox{\textwidth}{!}{%
\SmartTableNine{
Qwen3.5 Fine-Tuning;0.407056;0.458256;0.924866;10.576325;11.354451;1.155671;0.675353;0.416169;0.891837\\
Qwen3.5 Fine-Tuning + BioNER;0.454774;0.523334;0.924424;7.670661;10.151334;1.094759;0.308317;0.553081;0.890397\\
Qwen3.5 Fine-Tuning + RAG;0.337914;0.425323;0.905526;9.747804;10.782374;1.068232;0.493345;0.368356;0.878972\\
Qwen3.5 Fine-Tuning BioNER + RAG;0.282567;0.333187;0.873320;12.708475;11.303599;1.429540;0.574133;0.283183;0.808818\\
BioBART Fine-Tuning;0.543822;0.595321;0.942187;7.286685;10.283473;1.049745;0.612921;0.662893;0.934421\\
BioBART Fine-Tuning + BioNER;0.533472;0.584450;0.939945;6.085597;10.038004;1.001730;0.600545;0.630058;0.925454\\
BioBART Fine-Tuning + RAG;0.452012;0.485605;0.927750;6.302383;9.972648;1.403133;0.445173;0.406171;0.921938\\
BioBART Fine-Tuning + BioNER + RAG ;0.363366;0.379226;0.913315;5.930775;9.651242;2.063718;0.613802;0.292954;0.906036\\
}{
0.282567;0.333187;0.873320;5.930775;9.651242;1.001730;0.308317;0.283183;0.808818
}{
0.543822;0.595321;0.942187;12.708475;11.354451;2.063718;0.675353;0.662893;0.934421
}%
}
\end{table*}

\begin{table*}[!htp]
\centering
\caption{Mean scores for the best FT and few-shot strategies for each model}
\label{tab:mean_scores_best_ft_few_shot_each_model}
\resizebox{\textwidth}{!}{%
\SmartTableNine{
BioBART Fine-Tuning + BioNER ;0.533472;0.584450;0.939945;6.085597;10.038004;1.001730;0.600545;0.630058;0.925454\\
Qwen3.5 BioNER 0-Shot;0.358850;0.399018;0.918335;6.347393;9.384134;1.068674;0.602776;0.412674;0.894134\\
Qwen3.5 Fine-Tuning + BioNER ;0.454774;0.523334;0.924424;7.670661;10.151334;1.094759;0.308317;0.553081;0.890397\\
BioBART 0-Shot;0.194001;0.280669;0.844770;15.386576;11.901612;1.044979;0.300042;0;0.770808\\
}{
0.194001;0.280669;0.844770;6.085597;9.384134;1.001730;0.300042;0;0.770808
}{
0.533472;0.584450;0.939945;15.386576;11.901612;1.094759;0.602776;0.630058;0.925454
}%
}
\end{table*}

\subsection{Discussion}

The experimental results partially support the initial hypotheses. The BioNER strategy consistently improved readability and generally enhanced overall performance in both few-shot and fine-tuning settings. This suggests that explicitly providing biomedical entity information helps the models better identify important concepts while generating lay summaries.

In contrast, the RAG strategy did not consistently improve performance. Manual inspection showed that the retrieval system occasionally returned Wikipedia entries corresponding to terms with identical surface forms but different meanings, introducing irrelevant background information into the generation process. Although the prompt instructed the model to ignore unrelated retrieved content, the FENICE scores indicate that hallucinated information was still introduced in some summaries.

The combination of BioNER and RAG did not produce the expected improvements. Analysis revealed that several multi-word biomedical entities extracted by the BioNER system could not be matched by the Wikipedia API, resulting in missing or incomplete retrieved knowledge. Consequently, the potential benefits of retrieval were diminished, leading to lower overall performance.

The comparison between the two language models highlights the importance of domain-specific pretraining. Although Qwen demonstrated stronger few-shot capabilities, BioBART benefited substantially from fine-tuning, ultimately achieving the best overall results. This finding suggests that biomedical pretraining provides a stronger foundation for task-specific adaptation, whereas larger general-purpose language models can remain competitive in low-resource settings without additional training.

\subsection{Positive Impact}

This study shows that radiology reports can be made easier for patients to understand without removing the main clinical information. Lay summaries may help patients understand their results better and ask more useful questions during medical appointments. The findings also suggest that using biomedical entities can help the model focus on the most important parts of a report and explain them in clearer language.

Since the study uses small language models and LoRA fine-tuning, the proposed setup may also be practical for institutions with limited computing resources. At the same time, the RAG results show that adding external information is not always helpful. Wrong term matches or unrelated definitions can introduce information that is not supported by the report. This points to the need for more reliable medical knowledge sources and careful checking before these systems are used in practice.

These summaries should be used as an aid for patients and clinicians, not as a replacement for medical advice. Further testing with both patients and healthcare professionals is still needed before patient-facing use.

\section{Conclusion}

This study investigated the effects of BioNER- and RAG-based strategies on radiology report lay summarization under both few-shot inference and fine-tuning settings. The proposed approaches were evaluated using nine metrics covering relevance, readability, and factuality with equal weighting.

The experimental results show that the BioNER strategy consistently improved the baseline models, particularly in terms of readability, while also maintaining competitive relevance and factuality. In contrast, the RAG strategy did not consistently improve performance, and combining BioNER with RAG did not yield the expected gains. Overall, the findings partially support the initial hypotheses: BioNER proved to be an effective enhancement for lay summarization, whereas the effectiveness of RAG was limited by the quality of the retrieved knowledge.

These results demonstrate that providing explicit biomedical entity information is a simple yet effective approach for improving the readability of automatically generated lay summaries. Future work will focus on improving the retrieval component by exploring domain-specific knowledge bases, biomedical knowledge graphs, and more robust entity linking methods to reduce retrieval errors. In addition, investigating alternative BioNER models and retrieval strategies may further improve the quality and factual consistency of generated lay summaries.

\bibliographystyle{acl_natbib}
\bibliography{references}
\end{document}